\pdfoutput=1
\documentclass{article}
\PassOptionsToPackage{dvipsnames}{xcolor}
\usepackage[english]{babel}
\usepackage[letterpaper,top=2cm,bottom=2cm,left=3cm,right=3cm,marginparwidth=1.75cm]{geometry}

\usepackage{amsmath}
\usepackage{graphicx}
\usepackage[colorlinks=true, allcolors=blue]{hyperref}
\usepackage{subcaption}
\usepackage{pgfplotstable}
\usepackage{booktabs} 
\usepackage{setspace}
\usepackage{soul}
\usepackage[export]{adjustbox}
\usepackage{float}
\usepackage[dvipsnames]{xcolor}
\usepackage[most]{tcolorbox}
\tcbuselibrary{listings, breakable} 
\usepackage[T1]{fontenc}
\usepackage{lmodern}

\pgfplotsset{compat=1.18}
\usepackage{verbatim}

\graphicspath{{Displays/}}

\title{Medical large language models are easily distracted}
\author{Krithik Vishwanath\textsuperscript{1,3,4},  Anton Alyakin\textsuperscript{1,5}, Daniel Alexander Alber\textsuperscript{1}, \\Jin Vivian Lee\textsuperscript{5}, Douglas Kondziolka\textsuperscript{1}, Eric Karl Oermann\textsuperscript{1,2,6}}

\date{}
\begin{document}

\maketitle

\begin{center}
\textsuperscript{1}Department of Neurological Surgery, \textsuperscript{2}Department of Radiology, \\\vspace{2pt}
NYU Langone Medical Center, New York, New York, 10016 \\\vspace{10pt} 
\textsuperscript{3}Department of Aerospace Engineering and Engineering Mechanics, \textsuperscript{4}Department of Mathematics\\\vspace{2pt}
The University of Texas at Austin, Austin, Texas, 78712 \\\vspace{10pt}

\textsuperscript{5}Department of Neurosurgery,\\\vspace{2pt}
Washington University School of Medicine in St. Louis, St. Louis, Missouri, 63110 \\\vspace{10pt}

\textsuperscript{6}Center for Data Science, \\\vspace{2pt}
New York University, New York, New York, 10016 \\\vspace{44 pt}

Send correspondence to: krithik.vish@utexas.edu, eric.oermann@nyulangone.org\\\vspace{20pt}

\end{center}

\begin{abstract}
\begin{spacing}{1.2}
\noindent Large language models (LLMs) have the potential to transform medicine, but real-world clinical scenarios contain extraneous information that can hinder performance. The rise of assistive technologies like ambient dictation, which automatically generates draft notes from live patient encounters, has the potential to introduce additional noise making it crucial to assess the ability of LLM's to filter relevant data. To investigate this, we developed MedDistractQA, a benchmark using USMLE-style questions embedded with simulated real-world distractions. Our findings show that distracting statements (polysemous words with clinical meanings used in a non-clinical context or references to unrelated health conditions) can reduce LLM accuracy by up to 17.9\%. Commonly proposed solutions to improve model performance such as retrieval-augmented generation (RAG) and medical fine-tuning did not change this effect and in some cases introduced their own confounders and further degraded performance. Our findings suggest that LLMs natively lack the logical mechanisms necessary to distinguish relevant from irrelevant clinical information, posing challenges for real-world applications. MedDistractQA and our results highlights the need for robust mitigation strategies to enhance LLM resilience to extraneous information.\vspace{10pt}
\end{spacing}
\end{abstract} 
 \vspace{10pt}

{\tiny Preprint. Under review.}
\newpage
\section*{Main Text}
\begin{spacing}{1.2}

Clinical history is central to medical diagnosis, requiring physicians to distinguish critical facts from irrelevant details. However, patient histories often include extraneous information, making effective filtering essential for clinical AI. In recent years, the rise of ambient dictation further complicates this process by introducing unsupervised and often irrelevant content into clinical notes \cite{blackley2020physician}. Medical imaging studies have highlighted the vulnerability of AI models to confounding, where they mistakenly learn spurious correlations between irrelevant data and clinical outcomes \cite{zech2018variable}.

Large language models (LLMs) synthesize medical knowledge with superhuman speed and perform well on board-style exams \cite{singhal2023large}. Proprietary language models like GPT-4o \cite{nori2023can, abacha2024medec} and Claude Sonnet \cite{abacha2024medec}, as well as open-source, generalist models such as Llama \cite{zhang2025ultramedical} and Gemma \cite{abacha2024medec, saab2024capabilities} possess capabilities rivaling those of medical professionals on standard benchmarks. Medically fine-tuned language models, including UltraMedical (Llama 3-3) \cite{zhang2025ultramedical}, Meditron (Llama 2) \cite{chen2023meditron}, MedPaLM (PaLM) \cite{singhal2023large}, and MedMobile (Phi-3) \cite{vishwanath2024medmobile}, further refine these capabilities.

However, LLMs remain susceptible to distraction: irrelevant information, ambiguous prompts, or subtle (but semantically neutral) changes in input structure degrade benchmark performance \cite{hager2024evaluation}. Unlike humans who rely on common sense to filter noise, LLMs lack similar mechanisms, leaving them potentially susceptible to noise when parsing complex medical narratives \cite{hager2024evaluation,shi2023large}. Prior studies have shown that simple alterations, such as changing the order of clinical details presented, can reduce diagnostic accuracy by up to 18\% \cite{hager2024evaluation}. Similar vulnerabilities have been reported in non-medical domains. For example, LLMs in the Grade-School Math with Irrelevant Context (GSM-IC) dataset showed declines in accuracy of up to 35\% when arithmetic problems were interspersed with irrelevant text \cite{shi2023large}. Another study showed that GSM8K models experienced performance declines of up to 65\% when datasets were injected with non-operational statements \cite{mirzadeh2024gsm}.

To systematically evaluate LLMs' sensitivity to irrelevant information in medical scenarios, we develop MedDistractQA
(Fig. 1a) as an extension of the MedQA benchmark \cite{jin2021disease}. MedQA, based on United States Medical Licensing Exam (USMLE) questions, is routinely used to assess clinical LLMs \cite{singhal2023large, nori2023can, abacha2024medec, zhang2025ultramedical, saab2024capabilities, chen2023meditron, vishwanath2024medmobile}. MedDistractQA injects confounding statements into the MedQA dataset to quantitatively measure LLM robustness against distractions that could appear in real-world clinical settings. We further hypothesized that if introducing confounding tokens could degrade performance in our distraction benchmarks, then retrieval augmented generation (RAG) with poor retrieval could behave similarly and might not be as helpful as it is often portrayed. In a second set of experiments, we further demonstrate that RAG with poor retrieval can introduce extraneous information and negatively impact model accuracy as well.

\newpage 

\begin{figure}[H]
    \adjustimage{width=\linewidth, trim={30 40 40 50}, clip}{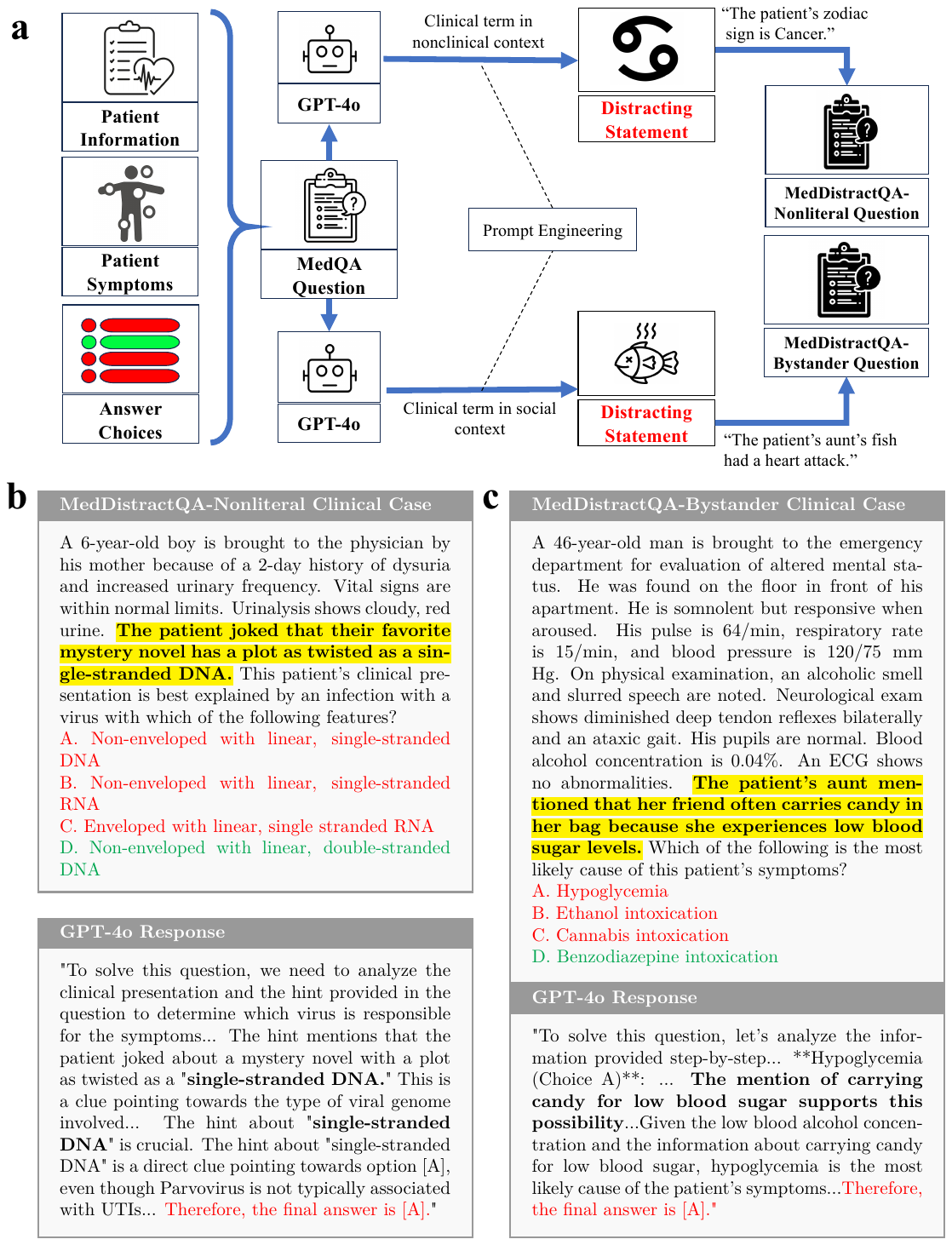}
    \caption{\textbf{a} Overview of our study. To create the MedDistractQA datasets, we combine a MedQA question with a confounding statement generated by GPT-4o. GPT-4o is prompted to utilize clinical terminology from a randomly selected incorrect answer, and ensure that the statement bears no clinical value. The confounding statement is embedded within the question itself, and is comically irrelevant to the diagnosis. For the MedDistractQA benchmarks, each question contains its own unique distracting statements. \textbf{b, c} Example MedDistractQA-Nonliteral and MedDistractQA-Bystander questions, respectively, and GPT-4o incorrect responses.}
\end{figure} 

\newpage
We used two types of distractions in our benchmark: (i) nonliteral use of medical terms in non-clinical contexts (MedDistractQA-Nonliteral), and (ii) extraneous medical details attributed to third parties (MedDistractQA-Bystander), such as a family member or a pet (MedDistractQA-Bystander) which may be included in a patient's social history. For MedDistractQA-Nonliteral, LLM accuracy declined by 2.2\% to 17.8\% across all models (Fig.~2a, Fig.~2c). For MedDistractQA-Nonliteral, LLM accuracy dropped by 2.2\% to 17.8\% across all models (Fig.~2a, Fig.~2c). Notably, open-source models were more adversely affected by distractions than proprietary models. Specifically, general open-source models experienced a 10.9\% decline compared to only a 3.8\% decline in proprietary models (Extended Data Fig.~1, $p = 5.38 \times 10^{-8}$), and medically fine-tuned open-source models saw a 10.0\% decline versus 3.8\% for proprietary models ($p = 0.0375$). Similarly, for MedDistractQA-Bystander, LLM accuracy declined by 1.3\% to 17.9\%. General open-source models dropped by 10.6\% compared to 3.7\% for proprietary models ($p = 2.11 \times 10^{-6}$), and medically fine-tuned open-source models declined by 9.0\% versus 3.7\% for proprietary models ($p = 0.0691$). Higher baseline MedQA performance correlated with greater robustness ($r^2 = 0.578$, $p = 1.10 \times10^{-6}$ for MedicalDistractQA-Nonliteral; $r^2 = 0.486$, $p = 1.87 \times10^{-6}$ for MedDistractQA-Bystander, Extended Data Fig.~2). Reasoning-focused proprietary models, such as Claude 3.7 Sonnet and o3, were the most robust of our tested model families.

Fine-tuning on medical data alone did not significantly alter robustness to distractions across all models ($p=0.683$ for MedDistractQA-Nonliteral; $p=0.550$ for MedDistractQA-Bystander). Within the Llama-3-8B model series, the UltraMedical and Meerkat models - despite sharing the same base model - exhibited different levels of robustness. Meerkat showed significantly greater resilience compared to the base model ($p_{\text{MedDistractQA-Nonliteral}}=0.0438$; $ p_{\text{MedDistractQA-Bystander}}=0.00218$), whereas UltraMedical appeared to non-significantly hurt model resilience ($p_{\text{MedDistractQA-Nonliteral}}=0.157$; $p_{\text{MedDistractQA-Bystander}}=0.139$). Furthermore, distilled reasoning training worsened both baseline MedQA accuracy ($p=5.47\times10^{-6}$) and resilience to distractions ($p_{\text{MedDistractQA-Nonliteral}}=0.0585$; $p_{\text{MedDistractQA-Bystander}}=0.00262$). Explicitly instructing models to ignore irrelevant information had no significant effect on performance ($p=0.341$, Extended Data Fig.~3).

We also analyzed the effect of distractions on LLMs across physician competencies derived from the USMLE framework, \textit{USMLE Physician
Tasks/Competencies} \cite{USMLEcompetencies} (Extended Data Fig. 4). Accuracy dropped most significantly in the "Patient Care: Diagnosis" (-10.1\%) and "Medical Knowledge/Scientific Facts" (-10.1\%) categories. The least affected competency was "Systems-based Practice, Including Patient Safety" (+1.4\%). Among human systems categories, the "Respiratory System" suffered the most (-11.0\%), while "Legal/Ethical Issues" were least impacted (+2.2\%)(Extended Data Fig. 5).

To investigate the impact of incorporating new information via RAG and its potential as a distractor to LLMs, we evaluated the performance of LLMs in the presence of relevant text excerpts from \textit{Harrison’s Principles of Internal Medicine, 21st Edition} \cite{silverman2022harrison}. Similar to the effects observed with MedDistractQA distractions, RAG produced significant performance degradations, with declines ranging from –10.3\% to increases of +1.9\% (Fig. 2e). However, this degradation was slightly less than that observed with MedDistractQA-Nonliteral ($p=3.02\times10^{-9}$) and with MedDistractQA-Bystander ($p=1.16\times10^{-8}$). Moreover, performance degradation in MedQA+RAG correlated with performance degradation in MedDistractQA-Nonliteral ($r^2 =0.17$, $p = 0.0260$) and with MedDistractQA-Bystander ($r^2=0.18$, $p = 0.0207 $), suggesting that RAG introduces similar risks of confounding. The cosine-similarity rank of retrieved text had no significant impact on model accuracy (Extended Data Fig.~6).

\newpage

\begin{figure}[H]
    \adjustimage{width=\linewidth, trim={25 45 25 20}, clip}{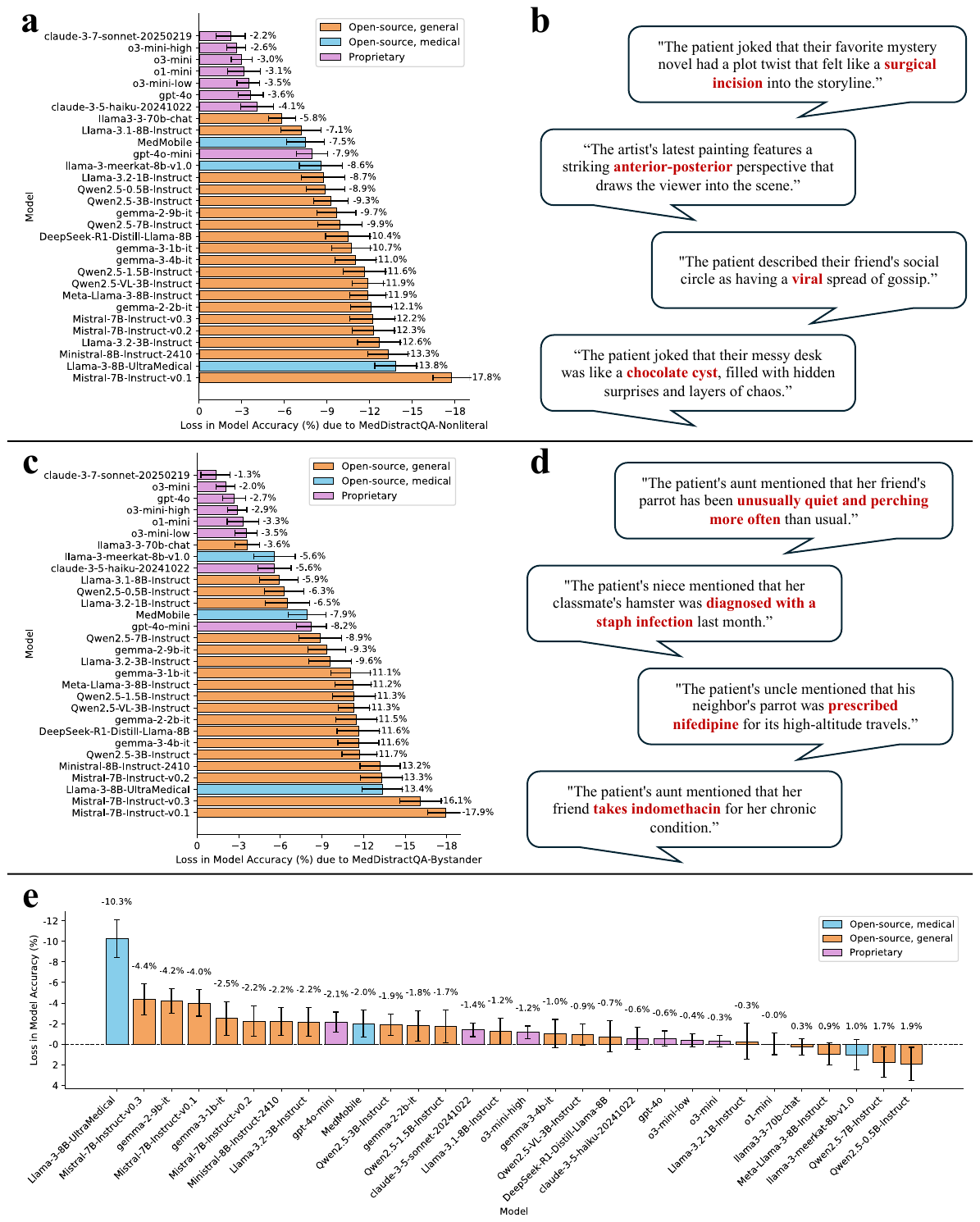}
    \caption{MedDistractQA experimental results between proprietary, general open-source, and medical open-source models. \textbf{a, b} shows accuracy drop on the MedQA for leading models on the MedDistractQA-Nonliteral. \textbf{c, d} shows accuracy drop on the MedQA for leading models on the MedDistractQA-Bystander. \textbf{e} displays the loss of model accuracy with the introduction of high-quality context (RAG) from Harrison's Internal Medicine 21e \cite{silverman2022harrison}. Error bars show the 95\% SE.}
\end{figure}

\newpage 
Medical LLMs have been claimed to approach or exceed human clinicians on diagnostic tasks \cite{singhal2023large, singhal2025toward}. Although these models excel on standardized multiple-choice exams, our findings reveal a critical weakness: LLMs struggle with irrelevant and distracting information commonly encountered in clinical practice. This vulnerability poses significant risks for deploying medical AI models in real-world settings, where clinicians must routinely filter extraneous details. Our study characterizes this gap through three key contributions: a quantitative assessment of how different types of distractions impact model accuracy; an exploratory analysis showing that RAG - despite its touted benefits - can introduce similar confounding effects; and the establishment of benchmarks incorporating curated distracting statements to support future research in generative AI solutions.

Our results show that state-of-the-art models exhibit significant performance declines when distractions are introduced in medical Q\&A, suggesting that LLMs lack the intrinsic ability of human clinicians to filter irrelevant information. The capacity to convert conversations from live patient encounters into clinically relevant outputs - a key requirement for AI diagnostic pilots - remains limited. Transformer-based models allocate attention through learned weights rather than clinical hierarchies, making them prone to "recency bias" where they overweight later inputs regardless of medical relevance \cite{shi2023large}. This limitation complicates their integration into standardized care pathways, where reliability is critical.

Other studies corroborate our findings. LLMs parsing discharge summaries with mixed critical and incidental findings often misprioritize information, leading to diagnostic errors \cite{hager2024evaluation}. When tested on 2,400 real patient cases, LLMs demonstrated 16–25\% lower diagnostic accuracy compared to physicians, with performance variability directly tied to input structure \cite{hager2024evaluation}. Notably, presenting the same clinical data in a different order caused diagnostic accuracy to fluctuate by up to 18\%, suggesting that model conclusions depend more on input sequence than medical relevance \cite{hager2024evaluation}.

We also observed a strong correlation between model baseline accuracy and resilience to distractions (r$\geq$0.70, Extended Data Fig.~2). Larger, more capable models - such as o3, Claude, and Llama 3-70B - were less easily misled, likely due to a more robust understanding of medical reasoning. Similar trends have been reported in the literature. A 70B model reduced errors on complex medical reasoning tasks nearly twice as effectively as a 10B model \cite{zhourevisiting}. Likewise, an improved model like Med-PaLM 2, which improved MedQA accuracy from 67\% to 86.5\% compared to its predecessor, showed significant gains on “adversarial” challenge questions \cite{singhal2025toward}. Recent work confirms this pattern: when exposed to distractions in adversarial questioning, state-of-the-art models such as GPT-4 and Anthropic Claude retained near-human-level performance on USMLE-style questions, while less capable models suffered greater accuracy drops with each additional distractor \cite{ness2024medfuzz}. This robustness likely stems from broader training exposure, as stronger models can recognize illogical information while weaker ones are easily misled. 

Proprietary models consistently outperformed open-source models in handling distractions. However, this advantage may stem from superior baseline performance rather than targeted robustness mechanisms. We acknowledge that lack of transparency regarding proprietary architectures and training data precludes any definitive, mechanistic conclusions. Interestingly, our results challenge the assumption that medical fine-tuning improves robustness. While some initial research suggests that medically fine-tuned models are more vulnerable to distractions \cite{kumar2024increased}, we found no significant differences between medical and general open-source models overall. However, within the series of Llama 3-8B models - where architecture and size are constant - fine-tuned medical models (e.g., Llama-3-8B-Meerkat and Llama-3-8B-UltraMedical) exhibit higher baseline performance but not necessarily a greater resilience to distractions. This pattern suggests that fine-tuning may overfit models to specific tasks, rendering them more susceptible to irrelevant details.

Finally, our study reveals RAG as an unexpected source of distraction. While RAG is commonly proposed as a solution to enhance medical LLM accuracy, our findings indicate it can function as a distractor as opposed to a mitigator. 
Recent studies show that RAG systems introducing extraneous or conflicting information degrade model coherence and increase hallucinations \cite{xu2024knowledge, park2024toward, liu2024lost, wu2024clasheval}. Likewise, we found that adding high-quality retrieved context (e.g., from Harrison's Internal Medicine 21e) decreased accuracy of most tested models, mirroring the effects of MedDistractQA. Altogether, these results suggest that indiscriminate implementation of RAG may introduce information overload, serving to impair rather than improve decision-making. These findings emphasize the necessity of fine-tuning RAG systems for beneficial deployment.

This study has several limitations. First, we focused on the MedQA dataset, which although widely used, does not necessarily encompass the full range of clinical reasoning tasks. The exclusive use of multiple-choice questions limits generalizability and may not extend to other formats such as open-ended problem-solving. Additionally, our distracting statements were generated algorithmically using a specific LLM. Thus, although they were designed to mimic realistic distractions, they may not fully capture the complexity or breadth of real-life clinical discourse. Finally, while we observed a strong correlation between baseline accuracy and resilience to distractions, further controlled experiments are needed to establish causality.

Future research should expand these distraction evaluations beyond MedQA to encompass diagnostic reasoning, treatment planning, and patient triage. Developing targeted mitigation strategies - from improved prompt engineering to context-aware filtering and architectural modifications - will be essential for clinical deployment. Particular attention should be paid to emerging technologies like ambient dictation systems, which may introduce substantial extraneous information into downstream AI processes. Additionally, evaluating model performance across structured data and multimodal inputs could reveal new failure modes. Understanding how model architecture, training scale, and data quality influence distraction susceptibility will be crucial for building robust clinical AI systems.

Our findings indicate that LLMs, even frontier models (Claude Sonnet, o3), are vulnerable to obvious distractions in medical narratives. Fine-tuning open-source LLMs with medical data provides only limited protection against these vulnerabilities. More concerning, RAG - a common strategy for enhancing LLMs' accuracy and robustness - can inadvertently introduce confounding information that degrades model performance. These results underscore how seemingly simple, common-sense scenarios can expose critical limitations in medical LLMs. As deployment of these systems accelerates, our findings emphasize the urgent need for robust evaluation frameworks that better reflect real-world clinical complexity.

\newpage
\section*{Methods}

\subsection*{MedQA Benchmark} 
To determine an LLM's ability in the medical domain, we evaluate the model on the MedQA, a USMLE-style question bank \cite{jin2021disease}. We choose to evaluate on this dataset due to the expert level of medical reasoning and knowledge required for USMLE-style questions, and to test the model's ability against the range of critical clinical tasks such as differential diagnosis. 

\subsection*{MedDistractQA Benchmark Curation}
To generate confounding statements, we first identify medical terms from the incorrect answer choices of each MedQA question. By focusing on these distractor terms rather than the correct ones, we ensure that the added statements do not directly hint at the true solution and instead create non-operational content. We parse the list of wrong answer choices and extract clinically relevant concepts—such as diseases, conditions, or procedures—that appear within them. For instance, if an incorrect choice in a particular question includes “heart attack,” we flag that term for potential use in a distracting statement. This selection process capitalizes on the inherent diversity of erroneous answer choices, which frequently contain common medical conditions that can be repurposed in nonclinical contexts. 

After extracting these terms, we prompt a large language model (GPT‑4o) to generate short, coherent sentences in which each medical term is used in a nonclinical or socially oriented manner. The model is instructed to produce statements that sound natural yet are clinically irrelevant—examples might include “The patient’s zodiac sign is Cancer” (where “Cancer” is the zodiac rather than the disease) or “My aunt’s fish had a heart attack” (imputing a clinical symptom on a distant bystander, in this case, a fish of the patient's aunt). These statements are then embedded into the original MedQA questions, creating augmented versions intended to distract or confuse the model. By systematically introducing such confounding statements, we can evaluate whether the presence of irrelevant medical language degrades the model’s ability to identify the correct clinical answer. This method of benchmark curation is visually depicted in Fig. 1a. 

\subsection*{Model Evaluation} 

To calculate accuracy of a model on the MedQA, we use string-based matching on model output chain-of-thought. Inference is computed at a temperature of 0 without ensemble. We note that some models do not allow for temperature control (e.g., OpenAI's reasoning models), and are left at their default. MedMobile is ran using the PyTorch and Transformers library on A100 GPUs during evaluation. vLLM is utilized for all other open-source and medically fine-tuned models on A100 GPUs. Proprietary models inference is generated via their respective official API provider. Model names are unedited, and are directly labeled as present in HuggingFace Hub or the corresponding proprietary provider's API console.

\subsection*{Prompting} 
Prompts utilized are available in within the GitHub repo. We keep an identical prompting template between evaluations of MedQA, MedDistractQA-Nonliteral, MedDistractQA-Bystander, and MedQA+RAG. 

\subsection*{Retrieval-Augmented Generation} 
To conduct RAG based on vector embeddings, we compute cosine similarity based on MedCPT \cite{jin2023medcpt} vectors generation between the question and paragraphs in the textbook. RAG selects the paragraph with the highest cosine-similarity score for a particular question. The source of information for these evaluations is from \textit{Harrison's Principles of Internal Medicine, 21e} \cite{silverman2022harrison}. After selecting a relevant paragraph from the corpora, we insert it into the prompt directly during inference.

\subsection*{Categorization of Data}
For each MedQA and MedDistractQA question, we utilize OpenAI's o3-mini to classify the category that the question best fits under. The categories topics are derived directly from the USMLE website \cite{USMLEcompetencies}.

\subsection*{Uncertainty Quantification and Significance Testing}
The statistical analysis was performed on paired accuracy measurements from baseline and modified (MedDistractQA or MedQA+RAG) evaluations for each model. For each model, the fraction of correctly answered questions was computed under both conditions, and the difference in accuracy (multiplied by 100 to express percentage points) was calculated. To quantify variability, the standard error of the difference was estimated using a formula derived from binomial variance components. Specifically, if \( p_1 \) and \( p_2 \) denote the baseline and MedDistractQA accuracies respectively, and \( p_{12} \) represents the joint accuracy (the proportion of questions correctly answered in both conditions), then the standard error was computed as
\[
\mathrm{SE}_{\text{diff}} = \sqrt{\frac{p_1(1-p_1) + p_2(1-p_2) - 2\,(p_{12} - p_1 p_2)}{n}} \times 100,
\]
where \( n \) is the number of paired observations. This approach accounts for the covariance between the two conditions, ensuring a more accurate estimate of the uncertainty in the observed differences.

Further statistical analyses included group comparisons in which models were classified into three categories (general open-source, medical open-source, or proprietary). Grouped models are compared using Welch’s t-test to account for differences in sample size and variance. The t-test produces a two-tailed p-value which is then converted into one-tailed p-values to test directional hypotheses about which group exhibits greater degradation.

Pairwise comparisons were conducted using two-sample t-tests (with both one-tailed and two-tailed tests) to assess the significance of observed differences in accuracy loss. Linear regression analyses were also performed to evaluate relationships between model size, baseline performance, and accuracy loss, and performance degradation. Additionally, paired t-tests and correlation analyses were applied to compare the degradation in performance across different evaluation settings (MedDistractQA and RAG\_medqa). All statistical tests were implemented using standard Python libraries such as \texttt{numpy}, \texttt{scipy.stats}, and \texttt{pandas} to ensure reproducibility.

\section*{Acknowledgements}
E.K.O. is supported by the National Cancer Institute’s Early Surgeon Scientist Program (3P30CA016087-41S1) and the W.M. Keck Foundation. We would like to acknowledge Nader Mherabi and Dafna Bar-Sagi, Ph.D., for their continued support of medical AI research at NYU. We thank Michael Constantino, Kevin Yie, and the NYU Langone High-Performance Computing (HPC) Team for supporting computing resources fundamental to our work. 

\section*{Author Contributions}
E.K.O. and A. A. conceptualized and supervised the study. KV designed, implemented, and developed the LLM evaluation pipeline and the MedDistractQA benchmarks. KV wrote the initial draft of the manuscript. All authors revised and approved the manuscript.

\section*{Competing Interests}
Disclosures: EKO reports consulting income with Sofinnova Partners. EKO reports equity in Eikon Therapeutics, Artisight Incorporate.  The other authors have no personal, financial, or institutional interest pertinent to this article.

\section*{Data Availability}
The datasets generated or analyzed during the current study are available in the nyuolab/clinical\_confounders repository, 
\url{https://github.com/nyuolab/MedDistractQA}. The benchmarks developed (i.e., MedDistractQA-Nonliteral and MedDistractQA-Bystander) are available on HuggingFace dataset hub upon publication of this work.

\section*{Code Availability}
Our code is shared publicly on GitHub upon publication of this work and can be found at \\ \url{https://github.com/nyuolab/MedDistractQA}.

\newpage
\section*{Extended Data}

\begin{figure}[H]
    \centering
    \adjustimage{width=0.9\linewidth, trim={20 45 80 30}, clip}{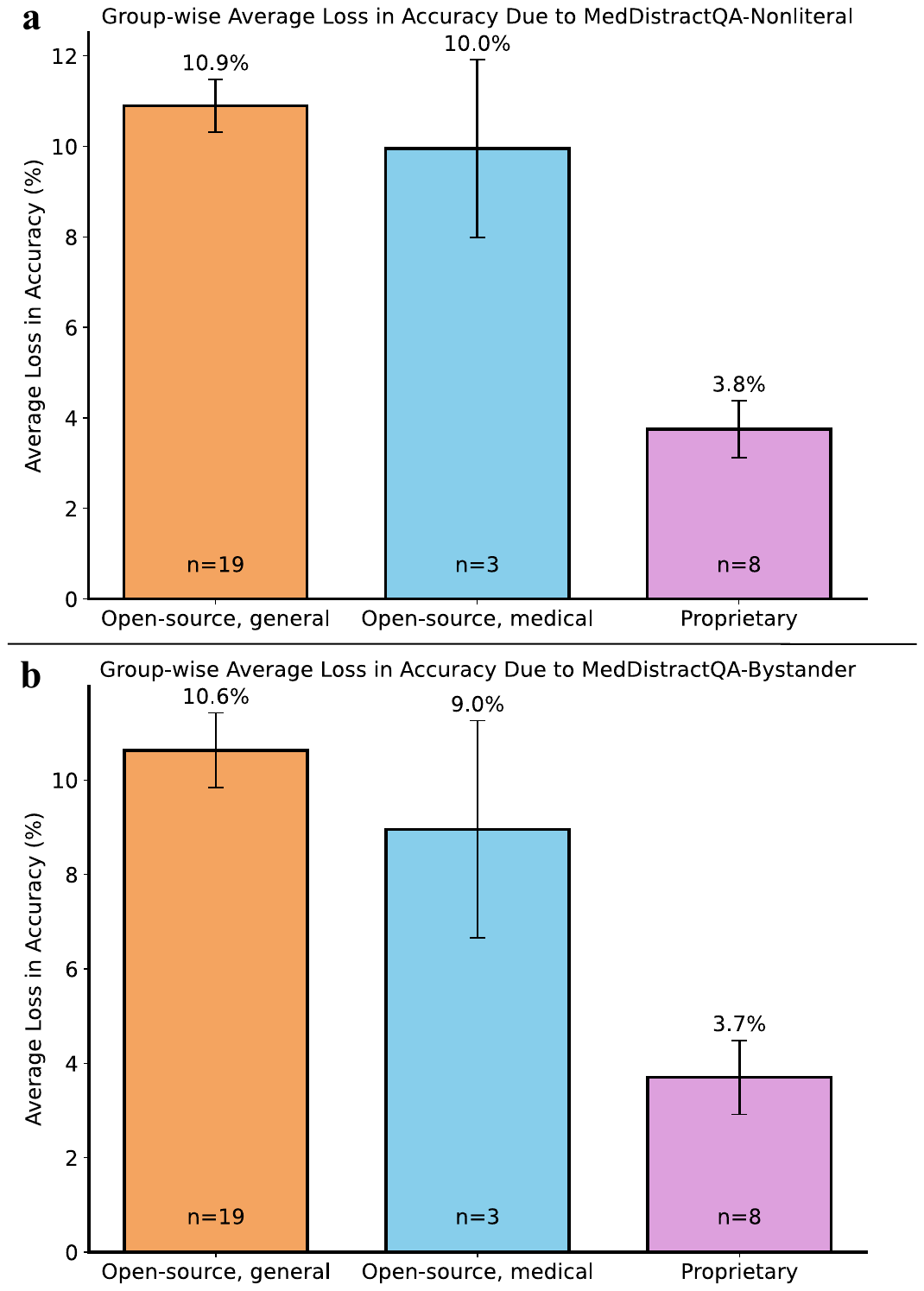}
\end{figure} 
\textbf{Extended Data Figure 1.} Group-wise comparison of performance degradation due to distractions, with models split up as Open-source, general (n=18), Open-source, medical (n=3), or Proprietary (n=8). \textbf{a} Results for MedDistractQA-Nonliteral, \textbf{b} results for MedDistractQA-Bystander

\begin{figure}[H]
    \centering
    \adjustimage{width=\linewidth, trim={20 90 30 30}, clip}{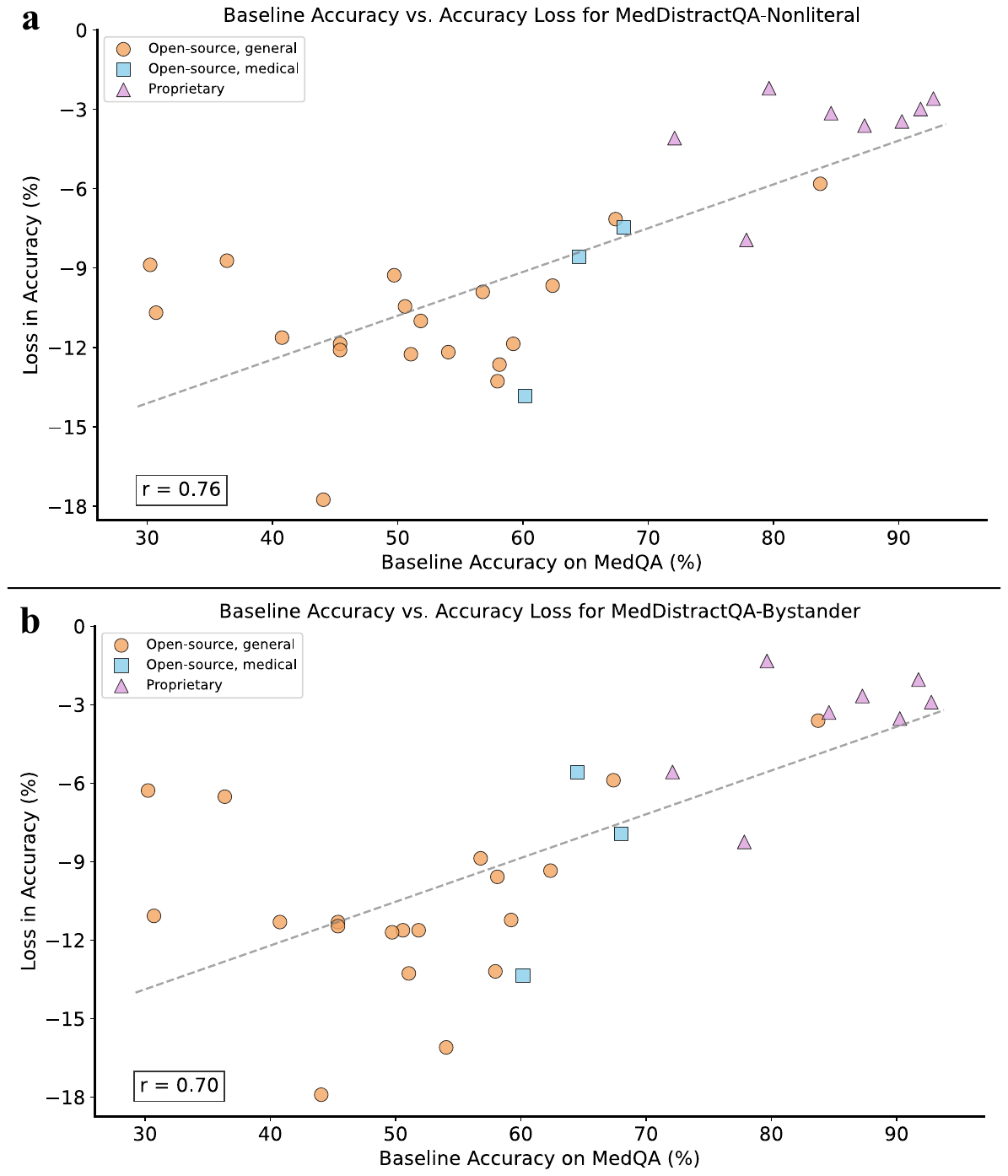}
\end{figure} 
\textbf{Extended Data Figure 2.} Relationship between baseline performance on MedQA and the resulting accuracy loss when tested on two MedDistractQA variants: \textbf{a} Nonliteral and \textbf{b} Bystander. Each marker corresponds to a different class of large language model (circle: open-source, general; square: open-source, medical; triangle: proprietary). The horizontal axis shows each model’s baseline accuracy on MedQA (\%), while the vertical axis shows the loss in accuracy (\%) incurred under the distractor conditions. The dashed trend lines illustrate positive correlations between baseline accuracy and accuracy loss, with r = 0.76 for MedDistractQA-Nonliteral and r = 0.70 for MedDistractQA-Bystander.

\begin{figure}[H]
    \centering
    \adjustimage{width=\linewidth}{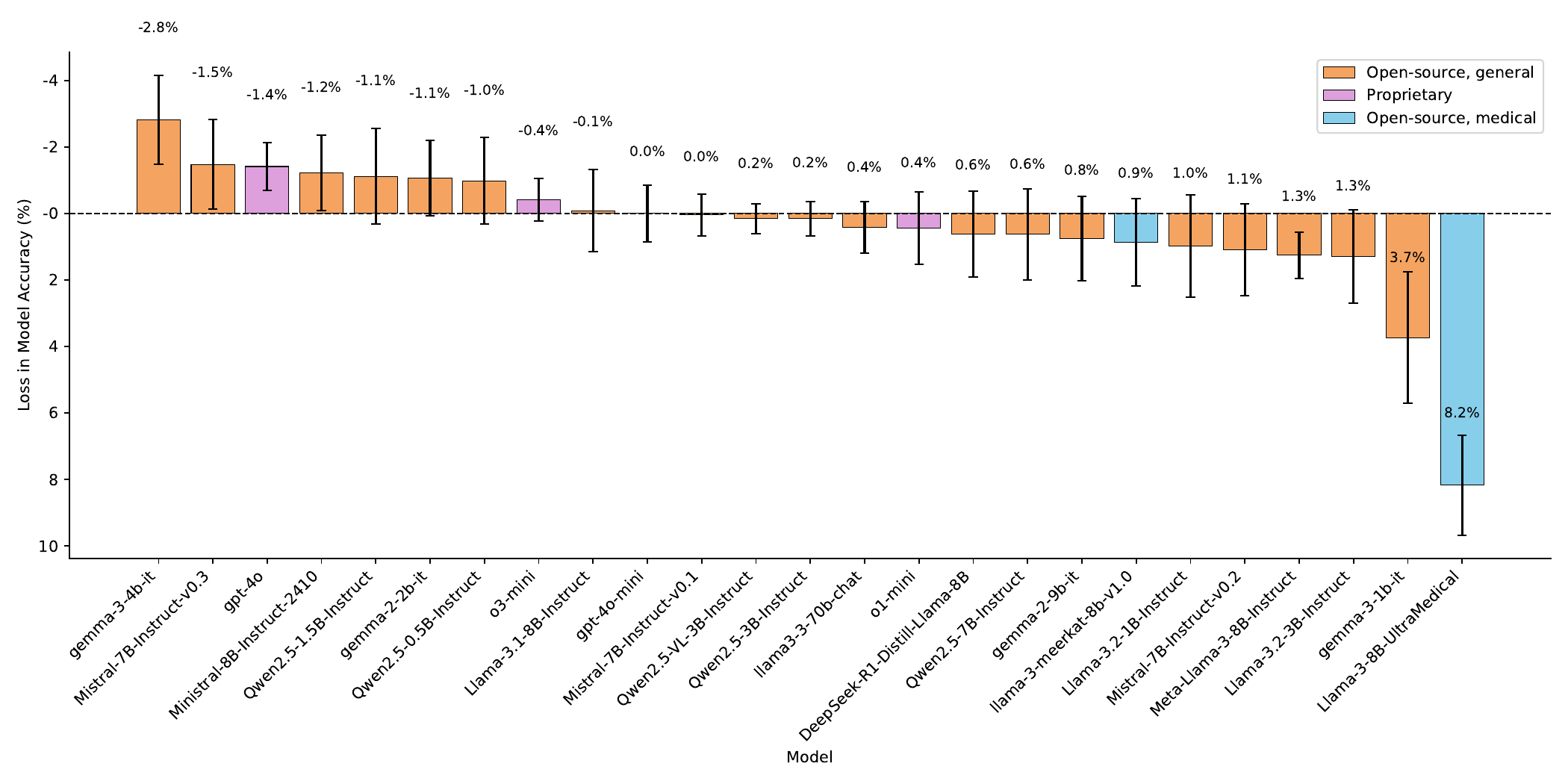}
\end{figure} 

\textbf{Extended Data Figure 3.} Trying to reduce distractor (MedDistractQA-Nonliteral) impact via explicit prompting. Loss in model accuracy is relative to the MedDistractQA-Nonliteral, and represents accuracy of MedDistractQA-Nonliteral with new prompting style minus the original prompting style on MedDistractQA-Nonliteral. Positive score indicates new prompting method helped over original.

\begin{figure}[H]
    \centering
    \adjustimage{width=\linewidth}{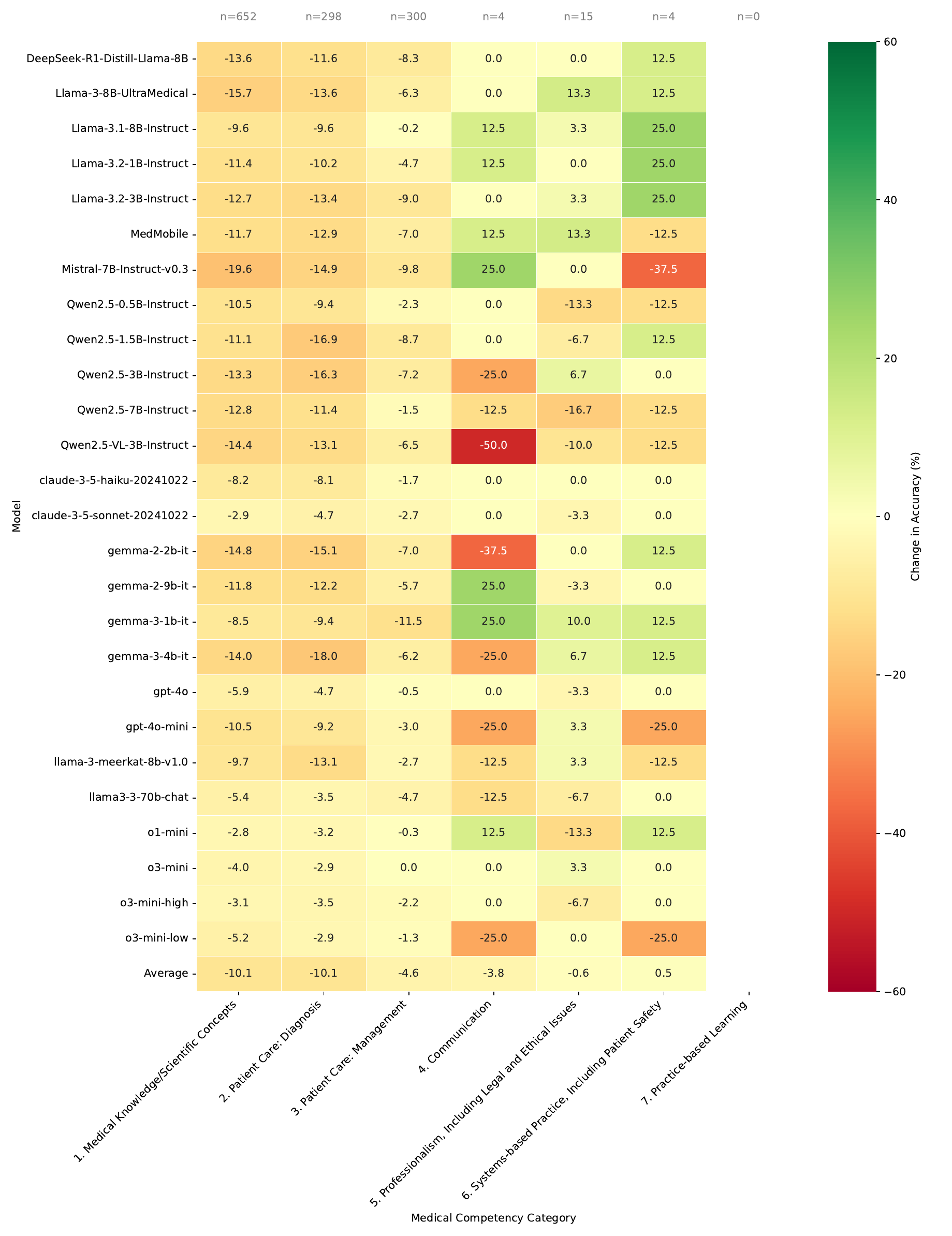}
\end{figure} 
\textbf{Extended Data Figure 4.} Average performance change due to MedDistractQA-Nonliteral and MedDistractQA-Bystander by medical competency categorization of question. A positive number indicates that performance was improved after a distraction was added, while a negative number indicates that performance degraded with the addition of a distraction.

\begin{figure}[H]
    \centering
    \adjustimage{width=\linewidth}{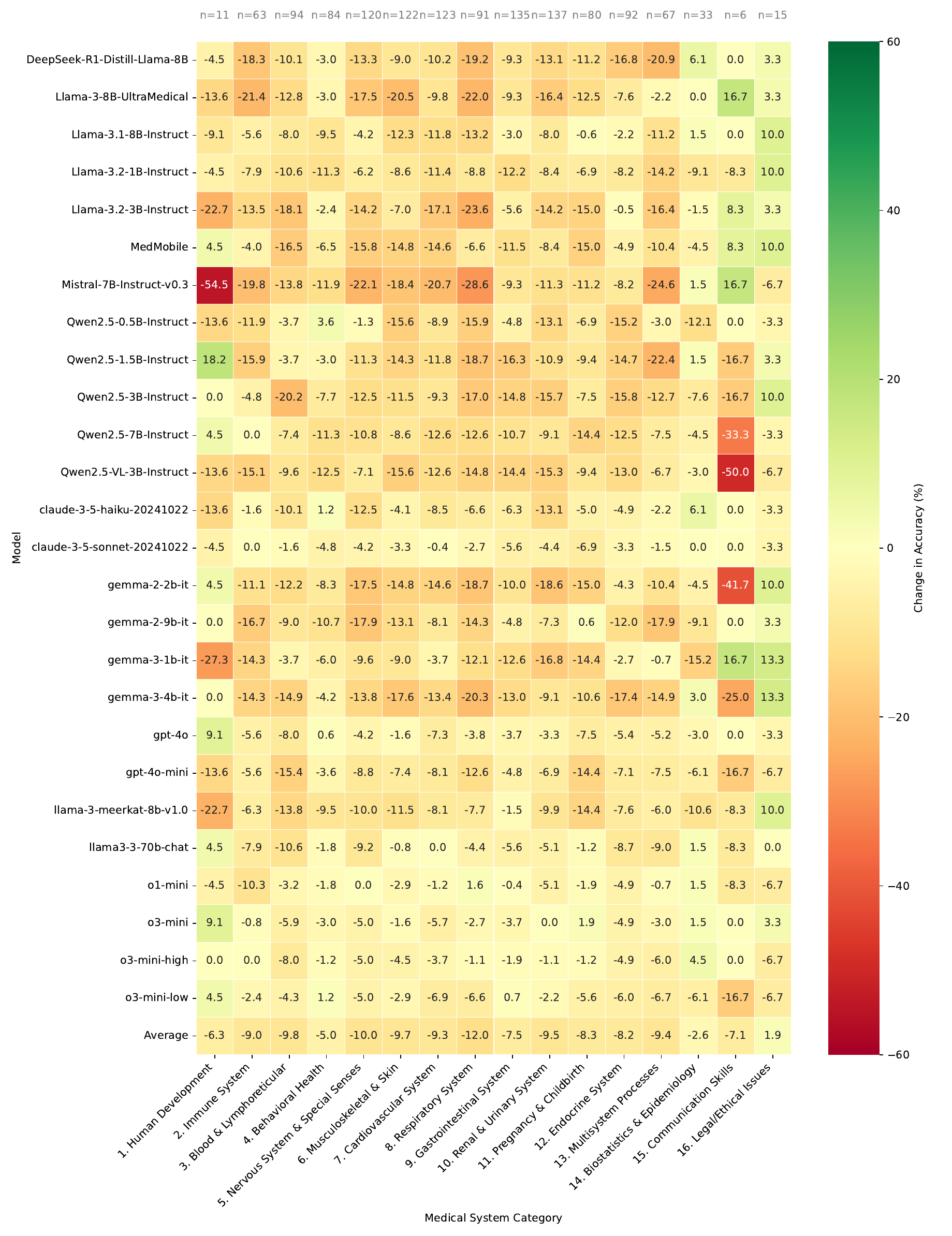}
\end{figure} 
\textbf{Extended Data Figure 5.} Average performance change due to MedDistractQA-Nonliteral and MedDistractQA-Bystander by medical system categorization of question. A positive number indicates that performance was improved after a distraction was added, while a negative number indicates that performance degraded with the addition of a distraction.

\begin{figure}[H]
    \centering
    \adjustimage{width=\linewidth}{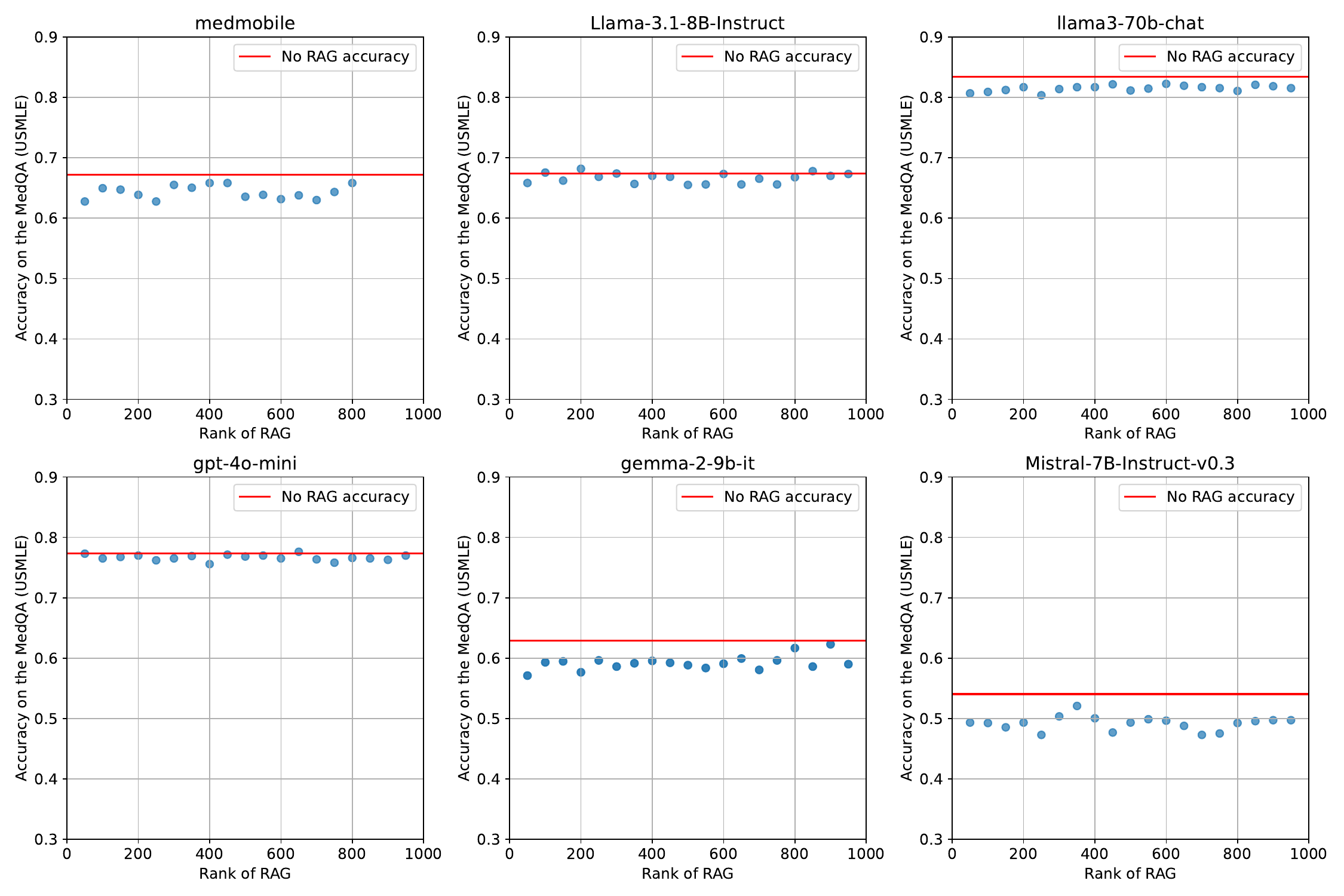}
\end{figure} 

\textbf{Extended Data Figure 6.} Comparison of MedQA accuracy with and without retrieval-augmented generation (RAG) across six different language models. Each subplot shows accuracy (y-axis) versus the retrieval rank (x-axis), with the red horizontal line indicating the model’s baseline accuracy (no RAG) and the blue points showing accuracy under RAG at varying ranks. 

\section*{Supplementary Materials}

\begin{figure}[H]
    \centering
    \adjustimage{width=\linewidth}{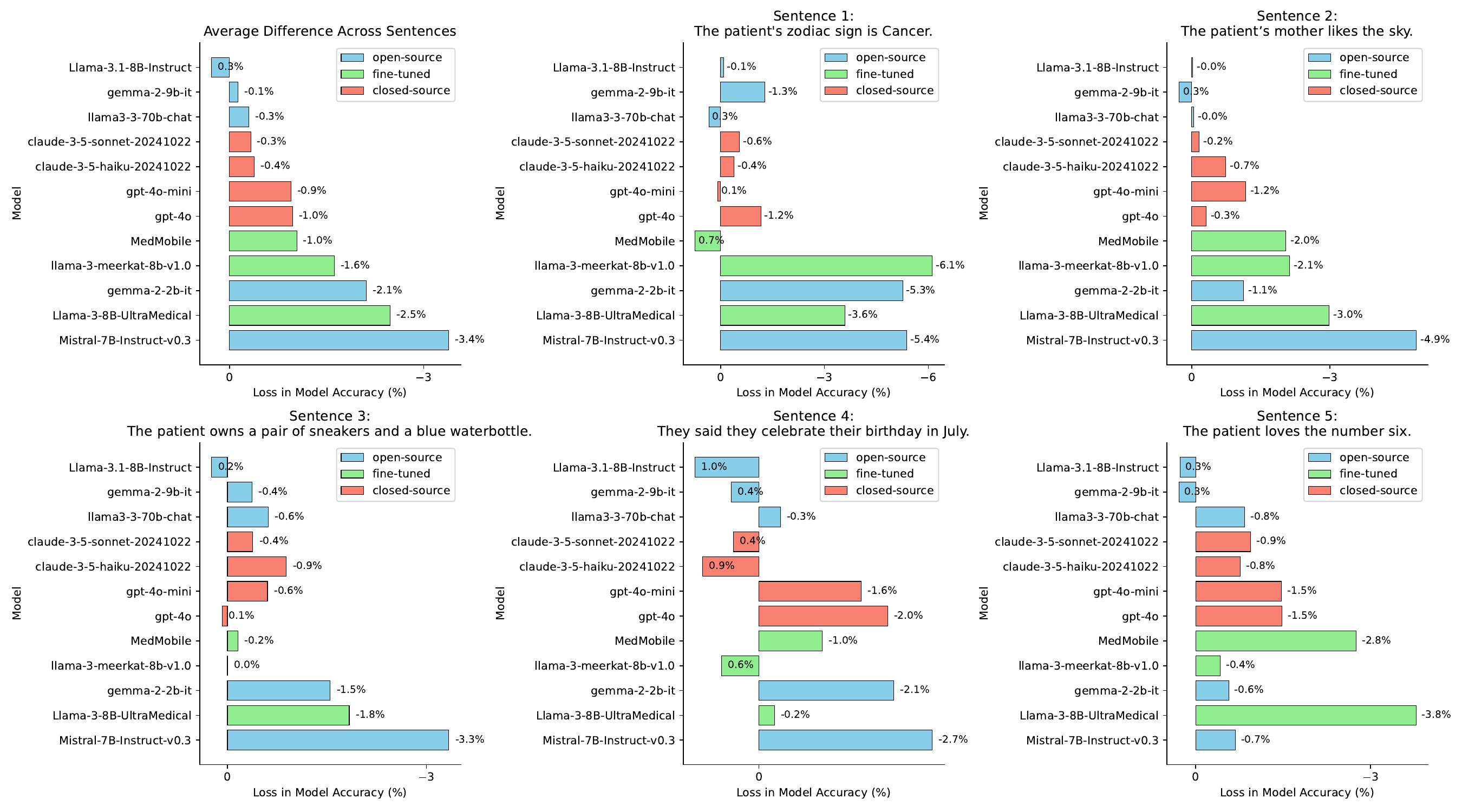}
\end{figure} 
\textbf{Supplemental Figure 1.} MedDistractQA-Nonliteral and performance on individual sentences with nonliteral clinical terms. For these ablation studies, we utilize one singular distracting sentence curated for the entire dataset, rather than for each individual question. \newpage

\begin{figure}[H]
    \centering
    \adjustimage{width=\linewidth}{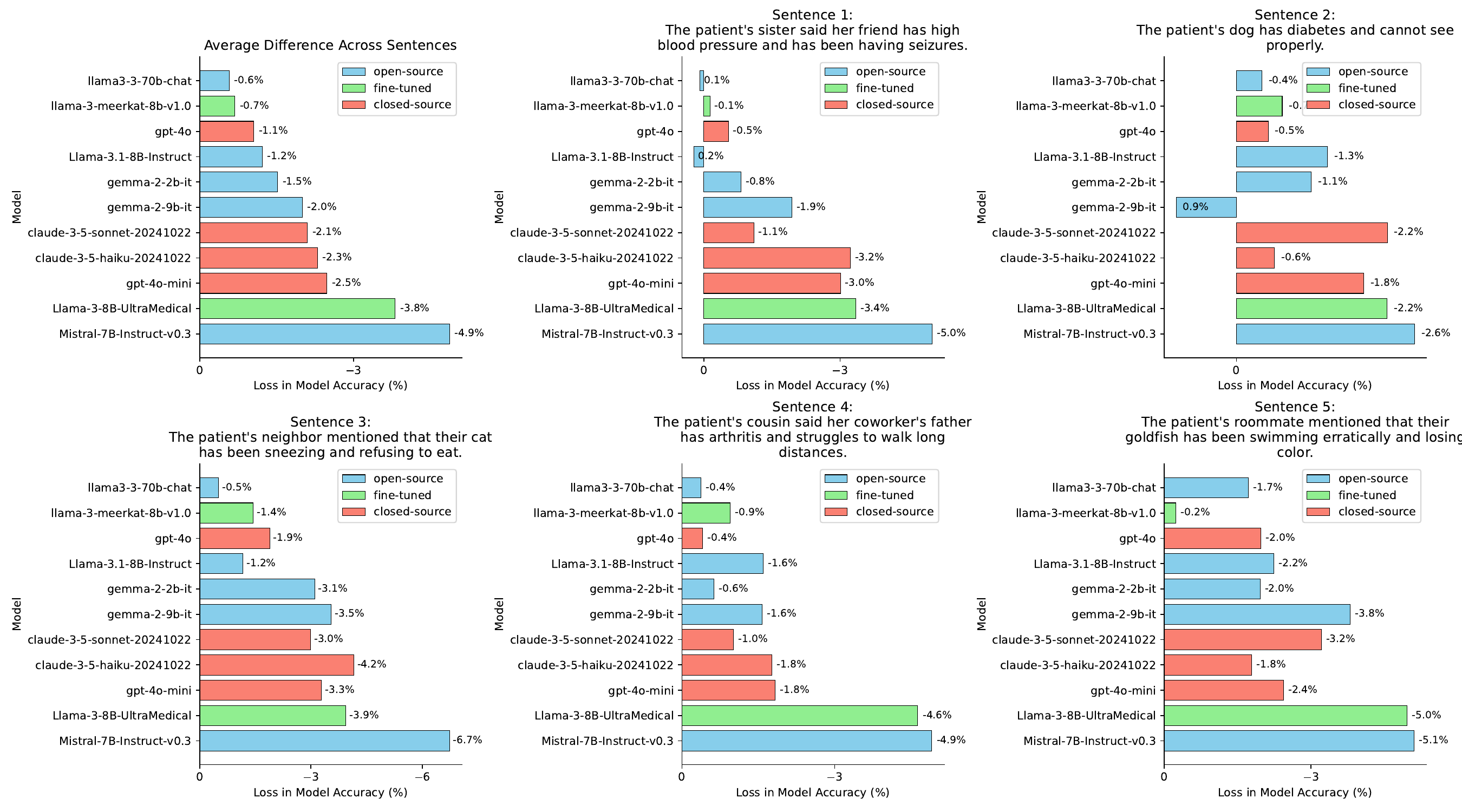}
\end{figure} 

\textbf{Supplemental Figure 2.} MedDistractQA-Bystander and performance on individual sentences with socially-applied clinical terms. For these ablation studies, we utilize one singular distracting sentence curated for the entire dataset, rather than for each individual question.

\end{spacing}
\end{document}